\documentclass[runningheads]{llncs}
\usepackage[T1]{fontenc}
\usepackage{xcolor}

\usepackage{graphicx}
\usepackage{soul}
\usepackage{booktabs,siunitx}
\usepackage{multirow}

\usepackage{color}
\usepackage{hyperref}
\begin{document}
\title{Structural Cycle GAN for Virtual Immunohistochemistry Staining of Gland Markers in the Colon}
\titlerunning{Structural Cycle GAN for Virtual Immunohistochemistry Staining}
\authorrunning{S. Dubey et al.}

\author{Shikha Dubey\inst{1,2}, Tushar Kataria\inst{1,2} \and Beatrice Knudsen\inst{3} \and
Shireen Y. Elhabian \inst{1,2,4} }

\institute{Kahlert School of Computing, University Of Utah \and
Scientific Computing and Imaging Institute, University of Utah \and Department of Pathology, University of Utah \and Corresponding author\\ \{\{shikha.d,tushar.kataria,shireen\}@sci, beatrice.knudsen@path\}.utah.edu }

\maketitle       
\begin{abstract}
With the advent of digital scanners and deep learning, diagnostic operations may move from a microscope to a desktop. Hematoxylin and Eosin (H\&E) staining is one of the most frequently used stains for disease analysis, diagnosis, and grading, but pathologists do need different immunohistochemical (IHC) stains to analyze specific structures or cells. Obtaining all of these stains (H\&E and different IHCs) on a single specimen is a tedious and time-consuming task. Consequently, virtual staining has emerged as an essential research direction. Here, we propose a novel generative model, Structural Cycle-GAN (SC-GAN), for synthesizing IHC stains from H\&E images, and vice versa. Our method expressly incorporates structural information in the form of edges (in addition to color data) and employs attention modules exclusively in the decoder of the proposed generator model. This integration enhances feature localization and preserves contextual information during the generation process. In addition, a structural loss is incorporated to ensure accurate structure alignment between the generated and input markers. To demonstrate the efficacy of the proposed model, experiments are conducted with two IHC markers emphasizing distinct structures of glands in the colon: the nucleus of epithelial cells (CDX2) and the cytoplasm (CK818). Quantitative metrics such as FID and SSIM are frequently used for the  analysis of generative models, but they do not correlate explicitly with higher-quality virtual staining results.  
Therefore, we propose two new quantitative metrics that correlate directly with the virtual staining specificity of IHC markers. 
\end{abstract}
\section{Introduction}
Histopathology image analysis has become a standard for diagnosing cancer, tracking remission, and treatment planning. Due to the large amount of data analysis required, the reduction in the number of pathologists in some domains  \cite{metter2019trends,jajosky2018fewer,robboy2020reevaluation} and the advent of digital scanners, automated analysis via deep learning methods is gaining importance in histopathology \cite{kataria2023automating,lu2023visual,wu2022recent}. Pathologists frequently use immunohistochemical (IHC) stain markers in addition to Hematoxylin and Eosin (H\&E) to highlight certain protein structures \cite{komura2023restaining} (not visible to the naked eye) that can help in identifying different objects essential for the analysis of disease progression, grading, and treatment planning. While some markers are accessible and time-efficient, others are not. Using conventional methods to stain a single slide with multiple stains in a pathology laboratory is time-consuming, laborious, and requires the use of distinct slides for analysis, thereby increasing the difficulty of the task. To address these issues, virtual staining offers a solution by generating all required stains on a single slide, significantly reducing labor-intensive and time-consuming aspects of conventional techniques.

Conditional-GAN (Pix2Pix) \cite{isola2017image} and Cycle-GAN \cite{zhu2017unpaired} are two pioneering generative adversarial networks (GAN) frameworks that have been frequently used in virtual staining \cite{rivenson2019virtual,rivenson2019phasestain,khan2023effect,xu2019gan} with many other variants proposed \cite{zingman2023comparative,shaban2019staingan,kang2021stainnet}. The performance of Pix2Pix based-method relies heavily on the data sampling procedure because this model is prone to hallucinations when patches of IHC are not precisely registered to their corresponding H\&E patches \cite{rivenson2019virtual,rivenson2019phasestain}. Registration of whole slide images (WSI) is a time-consuming and laborious process increasing the time to deployment and analysis. Although Cycle-GAN doesn't require registered images it doesn't explicitly focus on morphological relations between objects (cells, glands) of the input (histopathology) images. 
Here we propose a Structural Cycle-GAN (SC-GAN) model that alleviates the requirement of registering patches and focuses on the morphological relations. Using structural information \cite{xu2019gan} has previously shown promising results, but the proposed approach is novel in that it generates IHC stains based on structural information derived from the H\&E stain without relying on structural similarity loss. 
Incorporating structural information \cite{canny1986computational} provides explicit geometrical and morphological information to the model resulting in higher-quality virtually generated stains. 

Different IHC \cite{magaki2019introduction} stains highlight the nucleus, cell surface, or cytoplasm depending on the activating agent. To emphasize specific classes of cells, virtual staining models need to attend to a larger region of interest (cell environment). SC-GAN uses the attention module to factor in the dependence of pixel staining on its environment. Convolutional models have fixed ROIs around pixels of interest but the attention module can account for long-range dependencies \cite{raghu2021vision,chen2020dynamic}, which are beneficial in histopathology applications. 

Visual inspection by pathologists is the optimal method for evaluating the efficacy of any proposed virtual staining method, but it becomes impractical due to cost and pathologist disagreement \cite{arvaniti2018automated,eaden2001inter}. Quantitative evaluation metrics FID, SSIM, and PSNR \cite{liu2022bci,zingman2023comparative,borji2022pros} are heavily favored for the analysis of generative models. But in this study, we empirically observed that improvement in these metrics for virtual staining doesn't always correlate to better quality output. Therefore, two downstream task metrics are proposed, 1) Ratio of cell count in Stained IHC 
and 2) Dice Score of positive cells in Stained IHC, which correlates directly with virtual staining performance enhancement.  
The findings show that a lower FID score does not always imply a better quality stain, however the suggested downstream measures have significant correlations with stain quality. 

The key contributions of the present work can be summarized as follows:

\begin{enumerate}
\item Propose Structural Cycle-GAN (SC-GAN) model that employs explicit structural data (edges) as input to generate IHC stains from H\&E and vice versa. 
\item SC-GAN leverages the attention module to account for cell environmental dependencies and enforces structural consistency utilizing structural loss, thereby eliminating the need for registered patches.
\item Propose new evaluation metrics with a stronger correlation to enhanced virtual staining effectiveness and evaluate virtual staining results for two IHC markers of glands in the colon, CDX2 (specific to epithelial cell nuclei) and CK818 (cytoplasm), highlighting distinct cells. 
\end{enumerate}

\section{Methodology}
\vspace{-0.4in}
\begin{figure}
    \centering
    \includegraphics[scale=0.11]{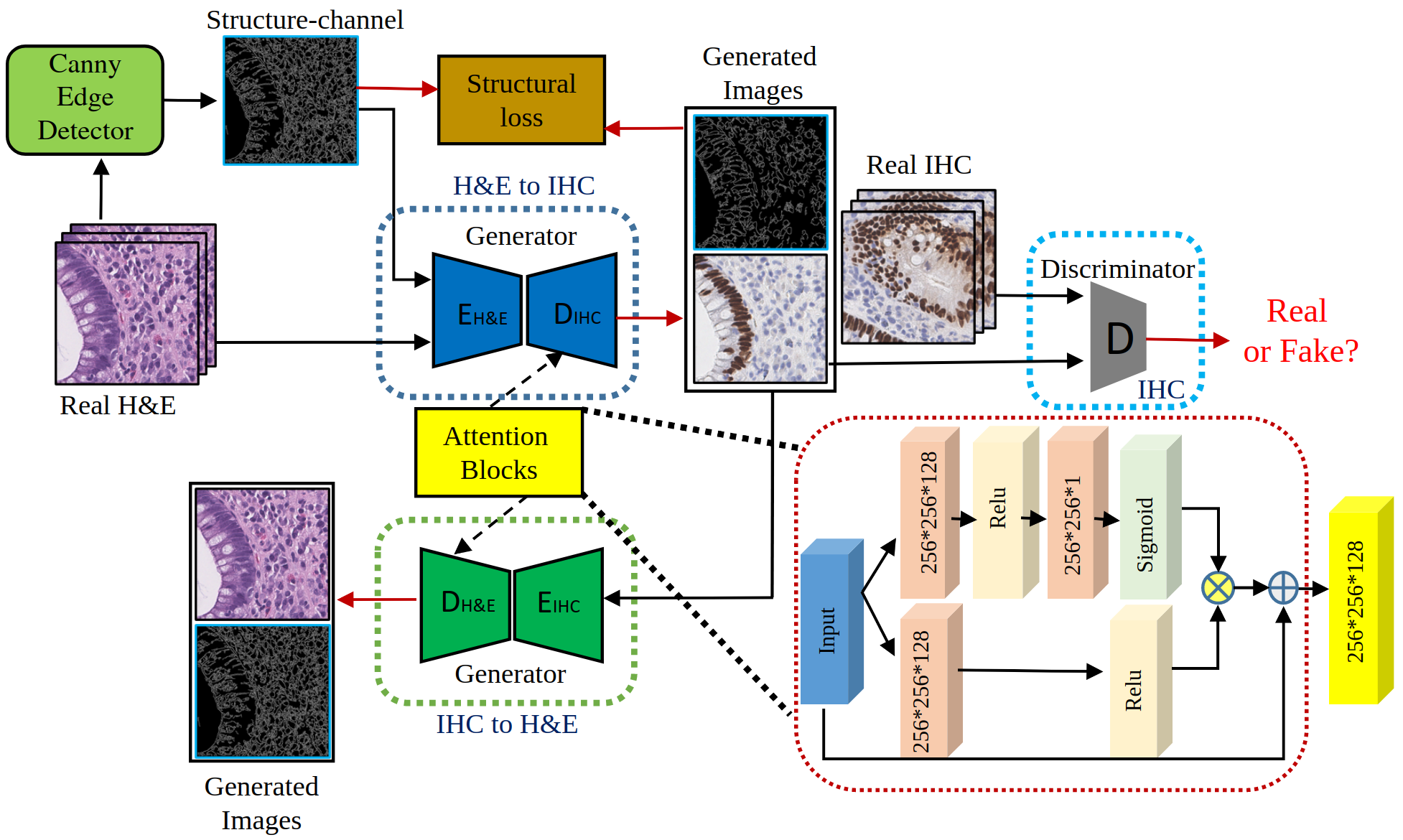}
    \vspace{-0.1in}
    \caption{\textbf{Block Diagram of the proposed Structural Cycle-GAN (SC-GAN)}: The Generator follows a ResNET-based architecture similar to \cite{xu2019gan}, with the addition of attention blocks in the decoder part. The proposed method does not necessitate the registration of real H\&E and IHC. The generation of H\&E stains from IHC stains follows the same architecture but incorporates the H\&E Discriminator, maintaining consistency in the overall model design and approach.
    }
    \label{fig:model}
    \vspace{-0.3in}
\end{figure}

\subsection{Structural Cycle-GAN (SC-GAN)}

The proposed model, SC-GAN (Figure \ref{fig:model}) is based on the Cycle-GAN framework \cite{zhu2017unpaired}, a prominent computer vision model for image-to-image translation tasks. The Cycle-GAN framework comprises two generators and two discriminators, incorporating cycle consistency loss to regularize GAN models in both domain \textit{A to B} and \textit{B to A} translations. Similarly, SC-GAN generators facilitate image translation between different domains {(H\&E to IHC} and vice versa), while the discriminators discern between real and generated images. The proposed approach introduces specific modifications {(mentioned in the following sections)} to the generator architecture while adhering to the fundamental principles of the Cycle-GAN framework.

\subsection{Structural Information}
Histopathology images frequently contain significant characteristics, such as cell boundaries, tissue structures, or distinct patterns. Histopathology imaging necessitates structure uniformity, as the structure of the stains must be maintained between stains. SC-GAN uses a canny-edge detector \cite{canny1986computational} to extract major structures/edges from the input stain and concatenate these with the brightfield color channel of the input patch before passing them through the Generator. By explicitly providing structure as input, SC-GAN conditions the generation process to take into account the cell's morphological as well as its texture properties. 
In addition to providing structural organization for cells, edges also serve as anchors to improve image generation. This contributes to the production of images with well-defined boundaries and structures, resulting in enhanced precision and aesthetically pleasing outcomes.
The generator synthesizes the target color-stained patches and their corresponding structures, which are then utilized as inputs for the second generator, as illustrated in Figure \ref{fig:model}.

\subsection{Structural Loss}
In addition to the losses used in Cycle-GAN \cite{zhu2017unpaired}, SC-GAN proposes a novel structural loss (SL) to improve the generation of IHC stains from H\&E stained images and vice versa. The SL is intended to ensure that structural information, such as margins and boundaries, is preserved in the generated stains. To calculate the SL, SC-GAN uses the Mean Squared Error (MSE) loss between the generated and the corresponding ground truth structural maps,  as shown in the equation below:
\begin{equation}
    \mathcal{L}_{SL} = MSE(I[S_{maps}]-G[S_{maps}]){,}
\end{equation}
where $I[S_{maps}]$ and $G[S_{maps}]$ stand for the structural maps (canny edges) of the input and generated stained images, respectively, and $\mathcal{L}_{SL}$ stands for the structural loss. In order to train the proposed model, the SL is combined with the other losses used in CycleGAN, such as adversarial loss (${L}_{adv}$), cycle-consistency forward (${L}_{cycle\_f}$) and backward losses (${L}_{cycle\_b}$), and identity loss ($\mathcal{L}_{I}$). The final loss is: 
\begin{equation}
    \mathcal{L} = \lambda_1 \mathcal{L}_{adv} + \lambda_2 (\mathcal{L}_{cycle\_f}+\mathcal{L}_{cycle\_b}) + \lambda_3 \mathcal{L}_{I} + \lambda_4 \mathcal{L}_{SL},
\end{equation}
 where $\lambda_1,\lambda_2, \lambda_3, \lambda_4$ are hyperparameters. 

\subsection{Attention-Enabled Decoder}
IHC staining not only depends on the cell type but also on the cell environment. To emphasize the role of the cell environment SC-GAN uses an attention module in the proposed generator. The attention block aims to enhance the generator's performance by selectively focusing on relevant cell environment details during the image translation process. The addition of attention in the decoder ensures that the generator concentrates on refining and enhancing the generated image, as opposed to modifying the encoded features during the translation process. The added attention module is depicted in Figure \ref{fig:model}.
\subsection{Additional Loss for Registered Data}

This study also evaluates SC-GAN on a registered dataset, where H\&E patches are paired with their corresponding registered IHC patches while training. SC-GAN is trained and evaluated with combination of registered and non-registered patches in varying amounts. 
To leverage the registered data more effectively, we introduce an additional supervision specifically for the registered data setting, utilizing the Mean Absolute Error (MAE) loss as given below:
\begin{equation}
    \mathcal{L}_{registered} = MAE(I_f-G_{f}),
\end{equation}
where $I_f$ and $G_{f}$ stand for the input and generated stains, and $\mathcal{L}_{registered}$ stands for the loss associated with the registered settings. The MAE loss measures the average absolute difference between the pixel intensities of generated markers and their corresponding ground truth markers. The experimental results for SC-GAN under registered settings are presented in Supplementary Table \ref{tab:registered-data}. 
\section{Results and Discussion}
\subsection{Dataset Details}
The dataset included H\&E WSIs from surveillance colonoscopies of 5 patients with active ulcerative colitis.
The slides were stained with H\&E using the automated clinical staining process and scanned on an Aperio AT2 slide scanner with a pixel resolution of 0.25 $\mu$m at 40x. After scanning, Leica Bond Rx autostainer was used for restaining by immunohistochemistry with antibodies reactive with CDX2 (caudal type homeobox-2) or CK8/18(cytokeratin-8/18).

We sampled patches from four patients, trained generation models (7,352 for CDX2 and 7,224 for CK818), and used data from the fifth patient for testing ($1,628$ for CDX2 and $1,749$ for CK818). Each patient's data consists of multiple tissue samples, ranging from $16$ to $24$. The extracted RGB patches are of shape $512\text{x}512$. The testing dataset contains registered H\&E and IHC patches which enables the development of two new metrics with a stronger correlation to the efficacy of the generative model in virtual staining. Further details on these metrics are provided in the section \ref{sec:eval_metric}.

\subsection{Model Architecture and Evaluation} \label{sec:eval_metric}
\noindent \textbf{Upsampling Method:} In contrast to the Conv2DTranspose method traditionally used for upsampling in generators \cite{zhu2017unpaired}, SC-GAN utilizes the UpSampling2D layer. This decision is based on empirical observations that demonstrate improved results with UpSampling2D in terms of preserving fine details and reducing artifacts. The discriminator is the same as proposed in the CycleGAN 
model, i.e., a patch discriminator. Hyperparameters are set to $\lambda_1=1.0,\lambda_2=10.0, \lambda_3=5.0,$ (the same as in the CycleGAN) and $\lambda_4=5.0$ (via experimentation).\footnote{We will release all the code related to the paper at a future date for public usage.}

\noindent \textbf{Proposed Evaluations:} SC-GAN reports the conventional evaluations such as FID and SSIM similar to other virtual staining works \cite{liu2022bci,zingman2023comparative}. However, our findings reveal that these metrics do not exhibit a significant correlation with improved virtual staining models (refer Figure \ref{fig:qual:cdx2} and Table \ref{tab:HE-to-ihc-eval}). Consequently, SC-GAN proposes the adoption of two new metrics that directly align with the performance of the generative model. 
\begin{itemize}
    \item \textbf{Ratio of cell count in Stained IHC}: Compares the number of cells highlighted in the virtually stained image with the real stained image utilizing DeepLIIF model \cite{ghahremani2022deepliif}. 
\begin{equation}
    \mathcal{R}_{count} =\left(\frac{|Gen_{cell}| - |GT_{cell}|}{|GT_{cell}|}\right)*100 ,
    \label{m_meq:counts}
\end{equation}
where $\mathcal{R}_{count}$ stands for ratio of cell count in Stained IHC with ground truth's cell counts. $|Gen_{cell}|$ and $|GT_{cell}|$ represent the number of generated and ground truth cells, respectively, calculated using the DeepLIIF model \cite{ghahremani2022deepliif}. A few examples of such segmented images from both stains, CDX2 and CK818 are shown in the Supplementary Figure \ref{fig:deepliif:example}. Additionally, DeepLIIF successfully segments the H\&E images for the IHC to H\&E evaluation (refer Supplementary Figure \ref{fig:Qual_HE} and Table \ref{tab:registered-data}).
\item \textbf{Dice Score of positive cells in Stained IHC}: Calculates the dice score and intersection of union (IOU) of the stained cells (positive cells) in the virtually generated image with respect to the real image. Both the metrics are evaluated by pixel-wise thresholding of brown color in generated and real IHC. Examples are depicted in Supplementary Figure \ref{fig:deepliif:example}.
\end{itemize}
\subsection{Results}
Table \ref{tab:HE-to-ihc-eval} presents the results of different models, including the proposed SC-GAN model, along with relevant comparisons, evaluated using various metrics. The conventional metrics, FID and SSIM, are included alongside the proposed metrics. The DeepLIIF model was utilized to segment and count the total number of cells, positive (IHC) cells, and negative (background) cells. 
The evaluation of cell counts was performed on a subset of the test dataset consisting of 250 registered patches, while the other metrics were calculated on full test dataset.
\begin{table}[!htb]
\centering
\caption{\textbf{Quantitative results for H\&E to IHC Translation.} EDAtt: Attention in both Encoder and Decode, DAtt: Attention in the decoder, St: Structure, SL: Structural Loss. Cell counting metrics (Total, Positive, and Negative) are calculated using equation \ref{m_meq:counts}, with values closer to zero indicating a more accurate generator. Qualitative results of pix2pix are not viable for cell counting analysis (refer to Figure \ref{fig:qual:cdx2}). IOU and DICE metrics are reported for IHC-positive cells.}
\scalebox{0.85}{
\begin{tabular}{l|c|c|c||c|c|c|c|c}
\hline
                     &         & \multicolumn{2}{c||}{Conventional} & \multicolumn{5}{c}{Proposed Metrics} \\ \hline
Markers              & \bf Models  & \bf FID  & \bf SSIM  & \bf IOU  &  \bf DICE & \bf Total & \bf Positive  & \bf Negative  \\ 
              &    &   &   &  & & \bf Cells & \bf Cells  & \bf Cells  \\ 
                     \hline
\multirow{6}{*}{CDX2}& Pix2Pix \cite{isola2017image} & \bf 6.58 & 18.54 &   -   &    -   &    -         &         -       &         -           \\ 
              & Base Cycle-GAN \cite{zhu2017unpaired} & 20.93& 38.23 & 51.79& 44.3 2 & 4.94       & 46.32          & -13.45              \\
              & Cycle-GAN w/ EDAtt& 13.34& 40.16 & 44.93& 46.25 & -9.57          &   -20.78         &   -5.63                \\
            & Cycle-GAN w/ DAtt& 12.74& \bf 40.64 & 45.91& 47.16 & -12.09          &   -11.65         &   -12.28                  \\
            & Cycle-GAN w/ St  & 15.00& 38.03 & 62.41& 50.83 &  -2.13          &       \textbf{2.90}    & -4.36                 \\ 
            & SC-GAN w/o SL    & 20.75& 35.53 & 56.86& 50.29&  -6.22         &    -21.73          &     7.31         \\ 
            & \bf \textit{SC-GAN(Proposed)}           & 14.05& 38.91 & \bf 63.35&  \textbf{51.73}     & \textbf{-0.08}  & -16.23            &          \textbf{1.63}                   \\ \hline
\multirow{6}{*}{CK818}& Pix2Pix \cite{isola2017image} &\bf 3.38& 28.93 &-   &  -     &     -        &      -          &   -                 \\ 
              & Base Cycle-GAN \cite{zhu2017unpaired} & 18.24& 27.03 & 23.36& 25.34 &   5.43     &   8.09        &      4.15         \\
              & Cycle-GAN w/ EDAtt& 15.02&  34.16 & 34.91& 33.19 & 5.83          &   10.87         &   3.40                 \\
            & Cycle-GAN w/ DAtt& 16.21& \bf 34.24 & \textbf{36.23}& \textbf{33.61} &    3.19       &   4.35           &   2.63                \\
            & Cycle-GAN w/ St  & 20.69& 32.35 & 35.86& 32.75 & -1.79          & \textbf{-2.04}             & -1.67                  \\ 
            & SC-GAN w/o SL    & 22.26& 32.22 & 14.25& 17.65 & 9.04          & -34.77             & 17.08                  \\ 
            & \bf \textit{SC-GAN(Proposed)}           & 15.32& 33.86 & 26.16& 27.66 &\textbf{0.22}         &   3.48             & \textbf{-1.35}                  \\ \hline
\end{tabular}}
\label{tab:HE-to-ihc-eval} 
\vspace{-0.2in}
\end{table}
 
\begin{figure}[!htb]
    \centering
    \includegraphics[scale=0.15]{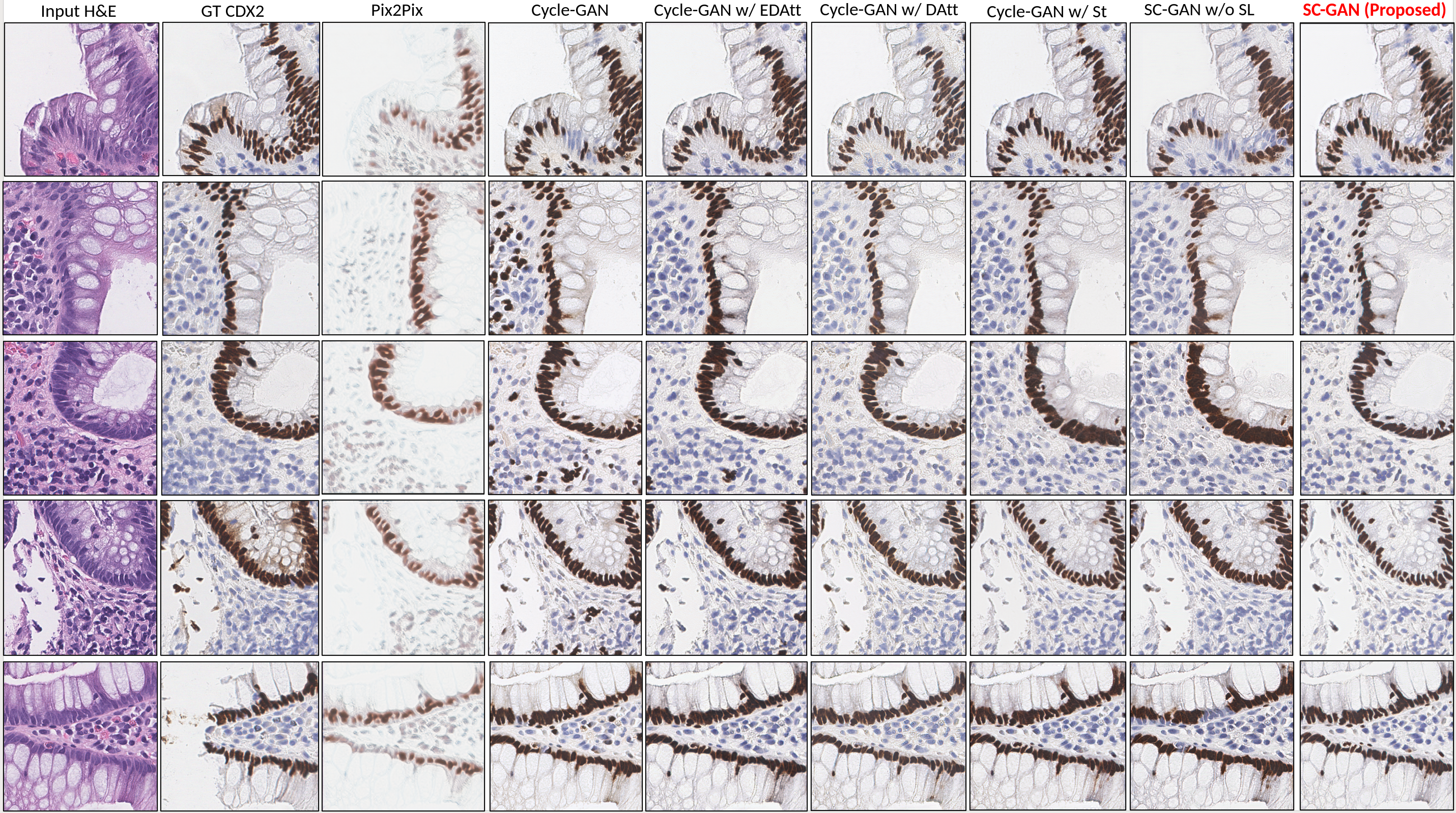}
    \vspace{-0.1in}
    \caption{\textbf{Qualitative results for generated CDX2 marker.} GT:Ground Truth, Gen:Generated, EDAtt: Attention in both Encoder and Decoder, DAtt: Attention in decoder, St: Structure, SL: Structural Loss. The proposed model, SC-GAN, performs better than the base Cycle-GAN by effectively suppressing false positives and accurately coloring the cells while preserving their structure. The final row showcases the effectiveness of virtual staining, where the given H\&E staining successfully reproduces the IHC stain, even though some information was lost during the original IHC staining process. Notably Cycle-GAN w/ DAtt performs better than Cycle-GAN w/ EDAtt model.}
    \label{fig:qual:cdx2}
\end{figure}

 From  Table \ref{tab:HE-to-ihc-eval}, we observe that a lower FID score is not always associated with a higher proposed metric that measures the number of valid IHC cells highlighted by the generative model. Conventional metric values indicate that the Pix2Pix model performs the best, despite producing qualitatively inferior results compared to other models (refer Figure \ref{fig:qual:cdx2}). Consequently, additional proposed metrics are required for a thorough assessment. Table \ref{tab:HE-to-ihc-eval} demonstrates that the cell IOU and DICE scores for the IHC images generated by the SC-GAN are higher. This indicates that the proposed model exhibits greater specificity. The proposed variations incorporating decoder attention, structural information, and SL consistently exhibit the lowest deviation from the true cell counts (total number of cells, positive and negative cells). These results demonstrate the superiority of the proposed model, SC-GAN, over the base Cycle-GAN and illustrate the significant impact of each module on the performance of virtual staining. Furthermore, the qualitative results depicted in Figure \ref{fig:qual:cdx2} align with these observations, as the proposed models consistently yield higher-quality results characterized by reduced false positives and increased true positives. 
 
SC-GAN has better performance while generating CDX2 stains compared to CK818 (refer Table \ref{tab:HE-to-ihc-eval}). As CDX2 is a nuclear marker, more precise structural information is required to generate an accurate stain. On the other hand, CK818 is a cytoplasmic (surface) stain, where the cell environment plays a more crucial role. Hence, attention-based models exhibit higher accuracy for CK818 stains, as they focus on capturing contextual information. This suggests that the importance of input features, whether structure or attention map, varies according to the type of stain being generated.

\begin{figure}
   \centering
   \includegraphics[scale=0.135]{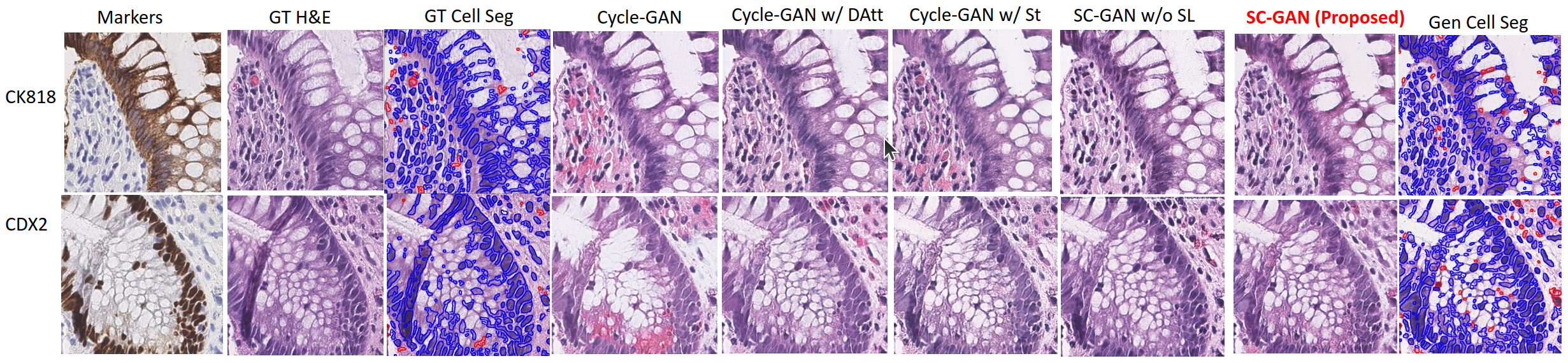}
   \label{fig:Qual_HE}
   \caption{\textbf{Qualitative results for generated H\&E from CDX2 and CK818 markers.} The proposed model, SC-GAN demonstrates superior performance compared to the base Cycle-GAN by accurately coloring the cells while preserving their structure. It also shows the segmentation of H\&E utilizing the DeepLIIF\cite{ghahremani2022deepliif} model.}
\end{figure}

The Supplementary Table \ref{tab:ihc-to-he} and Figure \ref{fig:Qual_HE} present the results for the IHC to H\&E translation. These results demonstrate that SC-GAN performs well not only for the H\&E to IHC translation but also for the reverse process. This observation is reinforced by the cell counting metric, which demonstrates the superior performance of SC-GAN in both translation directions. 
 
 \textbf{Registered data results: }
 In Supplementary Table \ref{tab:registered-data}, the influence of training the SC-GAN model with registered data is examined. The table reveals that incorporating registered data has minimal impact on the model's performance, indicating that SC-GAN can operate effectively without the need for registered data. Furthermore, it demonstrates that the translation capability of the SC-GAN model works in the feature space rather than relying solely on pixel-to-pixel translation. This characteristic enhances the robustness of the proposed SC-GAN model, making it suitable for virtual staining applications when there is limited availability of registered data.  
\section{Conclusion and Future Work}
We proposed a novel methodology, SC-GAN that combines structural information, attention, and a structural loss to enhance the generation of IHC markers from H\&E-stained images and vice versa. The proposed quantitative metrics focused on deviation in cell counts and dice score with respect to the real IHC images, shows direct correlation with virtual staining efficacy. Results highlight the need for different information in generative models based on stain properties (nuclear vs cytoplasmic, structure vs attention). SC-GAN outperforms in non-registered datasets, eliminating the need for laborious WSI image alignment. However, quantitative cell counting metrics rely on the real stain availability presents a limitation requiring further investigation. Additionally, future work can explore the potential benefits of integrating multi-scale structural information derived from wavelet \cite{graps1995introduction} or MIND  \cite{heinrich2013towards} features. The applicability of the proposed model to other IHC markers and its potential for generating marker-specific protocols are additional avenues for investigation. This could involve adapting a unified model architecture to effectively account for the unique characteristics associated with different stains. These can further improve the proposed methodology and increase its applicability for virtual staining in histopathology.

\section*{Acknowledgements} 
We thank the Department of Pathology, Scientific Computing and Imaging Institute, and the Kahlert School of Computing at the University of Utah for their support of this project. 
%
%

 \bibliographystyle{splncs04}
 \bibliography{paper_148}

\begin{thebibliography}{10}
\providecommand{\url}[1]{\texttt{#1}}
\providecommand{\urlprefix}{URL }
\providecommand{\doi}[1]{https://doi.org/#1}

\bibitem{arvaniti2018automated}
Arvaniti, E., Fricker, K.S., Moret, M., et~al.: Automated gleason grading of
  prostate cancer tissue microarrays via deep learning. Scientific reports
  \textbf{8}(1),  1--11 (2018)

\bibitem{borji2022pros}
Borji, A.: Pros and cons of gan evaluation measures: New developments. Computer
  Vision and Image Understanding  \textbf{215},  103329 (2022)

\bibitem{canny1986computational}
Canny, J.: A computational approach to edge detection. IEEE Transactions on
  pattern analysis and machine intelligence (6),  679--698 (1986)

\bibitem{chen2020dynamic}
Chen, Y., Dai, X., Liu, M., Chen, D., Yuan, L., Liu, Z.: Dynamic convolution:
  Attention over convolution kernels. In: Proceedings of the IEEE/CVF
  conference on computer vision and pattern recognition. pp. 11030--11039
  (2020)

\bibitem{eaden2001inter}
Eaden, J., Abrams, K., McKay, H., Denley, H., Mayberry, J.: Inter-observer
  variation between general and specialist gastrointestinal pathologists when
  grading dysplasia in ulcerative colitis. The Journal of Pathology: A Journal
  of the Pathological Society of Great Britain and Ireland  \textbf{194}(2),
  152--157 (2001)

\bibitem{ghahremani2022deepliif}
Ghahremani, P., Marino, J., Dodds, R., Nadeem, S.: Deepliif: An online platform
  for quantification of clinical pathology slides. In: Proceedings of the
  IEEE/CVF Conference on Computer Vision and Pattern Recognition. pp.
  21399--21405 (2022)

\bibitem{graps1995introduction}
Graps, A.: An introduction to wavelets. IEEE computational science and
  engineering  \textbf{2}(2),  50--61 (1995)

\bibitem{heinrich2013towards}
Heinrich, M.P., Jenkinson, M., Papie{\.z}, B.W., Brady, S.M., Schnabel, J.A.:
  Towards realtime multimodal fusion for image-guided interventions using
  self-similarities. In: Medical Image Computing and Computer-Assisted
  Intervention--MICCAI 2013: 16th International Conference, Nagoya, Japan,
  September 22-26, 2013, Proceedings, Part I 16. pp. 187--194. Springer (2013)

\bibitem{isola2017image}
Isola, P., Zhu, J.Y., Zhou, T., Efros, A.A.: Image-to-image translation with
  conditional adversarial networks. In: Proceedings of the IEEE conference on
  computer vision and pattern recognition. pp. 1125--1134 (2017)

\bibitem{jajosky2018fewer}
Jajosky, R.P., Jajosky, A.N., Kleven, D.T., Singh, G.: Fewer seniors from
  united states allopathic medical schools are filling pathology residency
  positions in the main residency match, 2008-2017. Human Pathology
  \textbf{73},  26--32 (2018)

\bibitem{kang2021stainnet}
Kang, H., Luo, D., Feng, W., Zeng, S., Quan, T., Hu, J., Liu, X.: Stainnet: a
  fast and robust stain normalization network. Frontiers in Medicine
  \textbf{8},  746307 (2021)

\bibitem{kataria2023automating}
Kataria, T., Rajamani, S., Ayubi, A.B., Bronner, M., Jedrzkiewicz, J., Knudsen,
  B., Elhabian, S.: Automating ground truth annotations for gland segmentation
  through immunohistochemistry  (2023)

\bibitem{khan2023effect}
Khan, U., Koivukoski, S., Valkonen, M., Latonen, L., Ruusuvuori, P.: The effect
  of neural network architecture on virtual h\&e staining: Systematic
  assessment of histological feasibility. Patterns  \textbf{4}(5) (2023)

\bibitem{komura2023restaining}
Komura, D., Onoyama, T., Shinbo, K., et~al.: Restaining-based annotation for
  cancer histology segmentation to overcome annotation-related limitations
  among pathologists. Patterns  \textbf{4}(2) (2023)

\bibitem{liu2022bci}
Liu, S., Zhu, C., Xu, F., Jia, X., Shi, Z., Jin, M.: Bci: Breast cancer
  immunohistochemical image generation through pyramid pix2pix. In: Proceedings
  of the IEEE/CVF Conference on Computer Vision and Pattern Recognition. pp.
  1815--1824 (2022)

\bibitem{lu2023visual}
Lu, M.Y., Chen, B., Zhang, A., Williamson, D.F., Chen, R.J., Ding, T., Le,
  L.P., Chuang, Y.S., Mahmood, F.: Visual language pretrained multiple instance
  zero-shot transfer for histopathology images. In: Proceedings of the IEEE/CVF
  Conference on Computer Vision and Pattern Recognition. pp. 19764--19775
  (2023)

\bibitem{magaki2019introduction}
Magaki, S., Hojat, S.A., Wei, B., So, A., Yong, W.H.: An introduction to the
  performance of immunohistochemistry. Biobanking: Methods and Protocols pp.
  289--298 (2019)

\bibitem{metter2019trends}
Metter, D.M., Colgan, T.J., Leung, S.T., Timmons, C.F., Park, J.Y.: Trends in
  the us and canadian pathologist workforces from 2007 to 2017. JAMA network
  open  \textbf{2}(5),  e194337--e194337 (2019)

\bibitem{raghu2021vision}
Raghu, M., Unterthiner, T., Kornblith, S., Zhang, C., Dosovitskiy, A.: Do
  vision transformers see like convolutional neural networks? Advances in
  Neural Information Processing Systems  \textbf{34},  12116--12128 (2021)

\bibitem{rivenson2019phasestain}
Rivenson, Y., Liu, T., Wei, Z., Zhang, Y., de~Haan, K., Ozcan, A.: Phasestain:
  the digital staining of label-free quantitative phase microscopy images using
  deep learning. Light: Science \& Applications  \textbf{8}(1), ~23 (2019)

\bibitem{rivenson2019virtual}
Rivenson, Y., Wang, H., Wei, Z., de~Haan, K., Zhang, Y., Wu, Y.,
  G{\"u}nayd{\i}n, H., Zuckerman, J.E., Chong, T., Sisk, A.E., et~al.: Virtual
  histological staining of unlabelled tissue-autofluorescence images via deep
  learning. Nature biomedical engineering  \textbf{3}(6),  466--477 (2019)

\bibitem{robboy2020reevaluation}
Robboy, S.J., Gross, D., Park, J.Y., Kittrie, E., Crawford, J.M., Johnson,
  R.L., Cohen, M.B., Karcher, D.S., Hoffman, R.D., Smith, A.T., et~al.:
  Reevaluation of the us pathologist workforce size. JAMA network open
  \textbf{3}(7),  e2010648--e2010648 (2020)

\bibitem{shaban2019staingan}
Shaban, M.T., Baur, C., Navab, N., Albarqouni, S.: Staingan: Stain style
  transfer for digital histological images. In: 2019 Ieee 16th international
  symposium on biomedical imaging (Isbi 2019). pp. 953--956. IEEE (2019)

\bibitem{wu2022recent}
Wu, Y., Cheng, M., Huang, S., Pei, Z., Zuo, Y., Liu, J., Yang, K., Zhu, Q.,
  Zhang, J., Hong, H., et~al.: Recent advances of deep learning for
  computational histopathology: principles and applications. Cancers
  \textbf{14}(5), ~1199 (2022)

\bibitem{xu2019gan}
Xu, Z., Huang, X., Moro, C.F., Boz{\'o}ky, B., Zhang, Q.: Gan-based virtual
  re-staining: a promising solution for whole slide image analysis. arXiv
  preprint arXiv:1901.04059  (2019)

\bibitem{zhu2017unpaired}
Zhu, J.Y., Park, T., Isola, P., Efros, A.A.: Unpaired image-to-image
  translation using cycle-consistent adversarial networks. In: Proceedings of
  the IEEE international conference on computer vision. pp. 2223--2232 (2017)

\bibitem{zingman2023comparative}
Zingman, I., Frayle, S., Tankoyeu, I., Sukhanov, S., Heinemann, F.: A
  comparative evaluation of image-to-image translation methods for stain
  transfer in histopathology. arXiv preprint arXiv:2303.17009  (2023)

\end{thebibliography}
 \newpage
 \section{Supplementary}
\begin{table}
\centering
\caption{\textbf{Quantitative results for IHC to H\&E Translation.} EDAtt: Attention in both Encoder and Decode, DAtt: Attention in the decoder, St: Structure, SL: Structural Loss. Cell counting metrics (Total Cells) are calculated using equation \ref{m_meq:counts}, with values closer to zero indicating a more accurate generator. SC-GAN performs well not only for the H\&E to IHC translation but also for the reverse process as shown by the results below.}
\scalebox{0.91}{
\begin{tabular}{l|c|c||c|c|c||cc}
\hline
                       & \multicolumn{3}{c|}{CDX2 to H\&E}  & \multicolumn{3}{c}{CK818 to H\&E} \\ \hline

                      & \multicolumn{2}{c||}{Conventional } & \multicolumn{1}{c|}{Proposed Metrics} & \multicolumn{2}{c||}{Conventional} & \multicolumn{1}{c}{Proposed Metrics}\\ \hline
             \bf Models  & \bf FID  & \bf SSIM  & \bf Total Cells & \bf FID  & \bf SSIM  & \bf Total Cells \\ 
                     \hline
 
               Base Cycle-GAN & 14.98& 39.33 &17.35    &  11.29& 24.40 &0.92            \\
               Cycle-GAN w/ EDAtt& 12.99&45.50 & 4.26 &14.1  & 28.08       & -4.99        \\
               Cycle-GAN w/ DAtt& 13.21& \textbf{45.53} & -4.99  &\textbf{10.45}  & 26.96       & -1.78         \\
             
             Cycle-GAN w/ St  & 14.41& 40.36 & 10.83  &  15.32 & \textbf{29.66}       & 2.07         \\ 
             SC-GAN w/o SL    & 14.47 & 43.51 & 7.87  &  11.72 & 25.79 & 3.18\\ 
             \bf \textit{SC-GAN(Proposed)}           & \textbf{12.9}& 44.05 &  \textbf{4.77} & 12.84& 28.13 & \textbf{-0.47}    \\ \hline
 
\end{tabular}}
\label{tab:ihc-to-he} 
\vspace{-0.3in}
\end{table}
\begin{table}
\centering
\caption{\textbf{Quantitative results for IHC to H\&E Translation for the registered data.} Cell counting metrics (Total, Positive, and Negative) are calculated using equation \ref{m_meq:counts}, with values closer to zero indicating a more accurate generator. Incorporating registered data has minimal impact on the model’s performance, indicating that SC-GAN can operate effectively without the need for registered data. IOU and DICE metrics are reported for IHC-positive cells. Non-Reg and Reg stand for non-registered and registered datasets for training.}
\scalebox{0.89}{
\begin{tabular}{l|c|c|c||c|c|c|c|c}
\hline
                     &         & \multicolumn{2}{c||}{Conventional} & \multicolumn{5}{c}{Proposed Metrics} \\ \hline
Markers              & \bf Models  & \bf FID  & \bf SSIM  & \bf IOU  &  \bf DICE & \bf Total & \bf Positive  & \bf Negative  \\ 
&    &   &   &  & & \bf Cells & \bf Cells  & \bf Cells  \\
                     \hline
\multirow{6}{*}{CDX2}&
               Reg SC-GAN & 19.71& 35.76 & 33.44& 38.54 & -16.56      &   -16.48        &   7.23            \\
            & Reg+10\% Reg SC-GAN& 14.6& \textbf{40.56} & 56.76& 48.22 & -1.08         &      \textbf{5.25}    &   -3.88                 \\
            & Non-Reg+30\% Reg SC-GAN  & \textbf{13.53}& 39.70 & 41.72& 43.64 &  -8.49          & -33.45  &     2.58             \\ 
            & Non-Reg+50\% Reg SC-GAN    & 27.42& 35.22 & 47.78& 46.39&      -11.46     &     12.27         &       -21.98     \\
            & Non-Reg+80\% Reg SC-GAN    &19.13 & 40.31 & 37.97& 40.6&  -15.13       &     -17.98         &        -13.87     \\
           & \bf \textit{SC-GAN (Non-Reg)}           & 14.05& 38.91 & \bf 63.35&  \textbf{51.73}     & \textbf{-0.08}  & -16.23            &          \textbf{1.63}                   \\ \hline
\multirow{6}{*}{CK818}
              & Reg SC-GAN & 23.50& 32.71 & 13.48&17.08 & 8.99       & -20.55          &    20.23          \\
            & Non-Reg+10\% Reg SC-GAN& 18.16& \textbf{36.26}& \textbf{33.86}& \textbf{33.73 }& 8.23       &    -18.13       &      3.45              \\
            & Non-Reg+30\% Reg SC-GAN  & 25.51& 34.06 & 25.29& 25.7 &      7.99     &    8.09   &    4.15             \\ 
            & Non-Reg+50\% Reg SC-GAN    & 23.77& 30.85 & 13.14&16.49&   20.09      &       3.53   &   33.80           \\
            & Non-Reg+80\% Reg SC-GAN    & 19.62& 35.55& 20.92& 24.03&       6.17    &   -15.33         &        16.53      \\
              & \bf \textit{SC-GAN(Non-Reg)}       &    15.32& 33.86 & 26.16& 27.66 & \textbf{0.22}         &   \textbf{3.48}             & \textbf{-1.35}                  \\ \hline
\end{tabular}}
\label{tab:registered-data} 
\end{table}

\begin{figure}[!htb]
    \centering
    \includegraphics[scale=0.15]{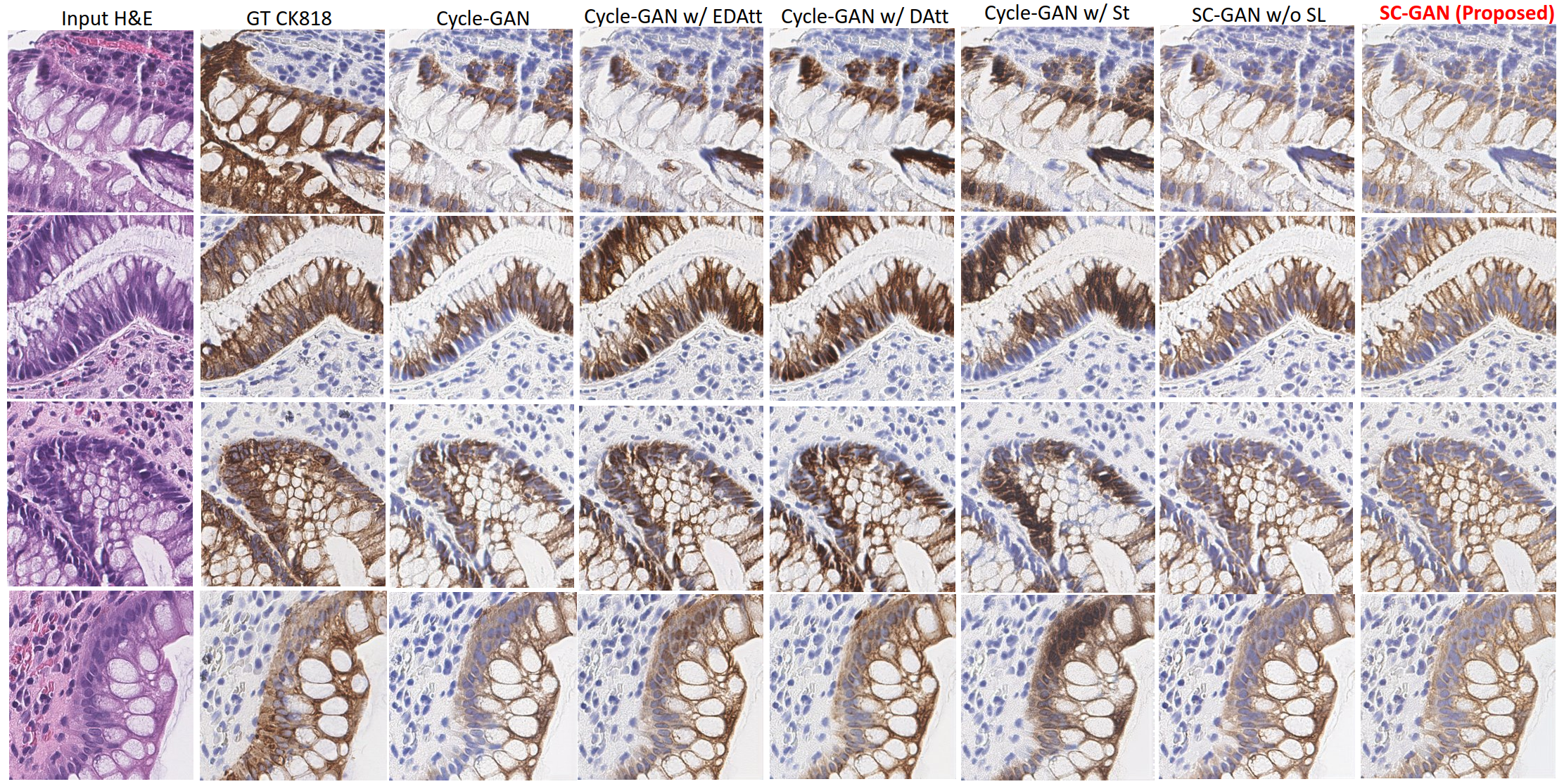}
    \caption{\textbf{Qualitative results for generated CK818 marker.} GT: Ground Truth, Gen: Generated, EDAtt: Attention in both Encoder and Decoder, DAtt: Attention in decoder, St: Structure, SL: Structural Loss. The superior performance of attention models demonstrates that CK818 is dependent on the cell environment for better results.
    Although lower brown color intensities lead to lower IOU and DICE scores for CK818 virtual markers (Table \ref{tab:ihc-to-he}), the impact on cell counting metrics is minimal, as evidenced by the quality of the threshold mask and segmented mask illustrated in Supplementary Figure \ref{fig:deepliif:example}. 
   }
    \label{fig:qual:ck818}
    \vspace{-.1in}
\end{figure}
\begin{figure}
    \centering
    \includegraphics[scale=0.15]{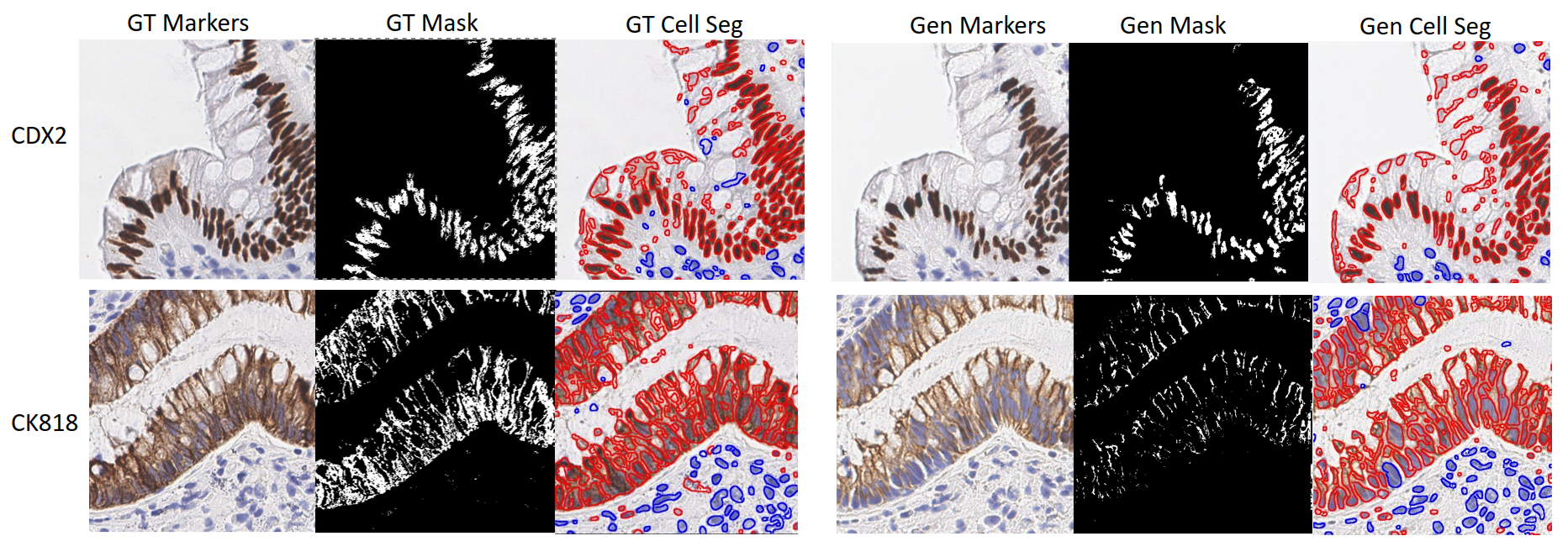}
    \caption{\textbf{Masking and Segmentation examples on CDX2 and CK818 markers.} GT: Ground Truth, Seg: Segmentation, Gen: Generated. GT and Gen Masks are created by pixel-wise thresholding of brown color, IOU and DICE for proposed metrics are calculated using both of these masks. GT and Gen Cell Seg are generated using the DeepLIIF\cite{ghahremani2022deepliif} model, which are utilized for proposed cell counting metrics.}
    \label{fig:deepliif:example}
\vspace{-0.03in}
\end{figure}

\end{document}